\documentclass[a4paper,twoside]{article}

\usepackage{epsfig}
\usepackage{subcaption}
\usepackage{calc}
\usepackage{amssymb}
\usepackage{amstext}
\usepackage{amsmath}
\usepackage{amsthm}
\usepackage{multicol}
\usepackage{pslatex}
\usepackage{apalike}
\usepackage[ruled,vlined,linesnumbered]{algorithm2e}
\DontPrintSemicolon
\SetAlgoLined
\usepackage[bottom]{footmisc}
\usepackage{url}
\usepackage{booktabs}
\usepackage{multirow}
\usepackage{array}
\SetKwInput{KwIn}{Input}
\SetKwInput{KwOut}{Output}
\SetKw{Continue}{continue}
\SetKwFunction{ClassifyMovement}{ClassifyMovement}
\SetKwFunction{GenerateCaption}{GenerateCaption}
\SetKwFunction{IsCached}{IsCached}
\SetKwFunction{GetCachedAudio}{GetCachedAudio}
\SetKwFunction{TextToSpeech}{TextToSpeech}
\SetKwFunction{CacheAudio}{CacheAudio}
\SetKwFunction{ParseResponse}{ParseResponse}
\SetKwFunction{GenerateSFX}{GenerateSFX}
\SetKwFunction{ShouldThrottle}{ShouldThrottle}
\SetKwFunction{SendToClient}{SendToClient}
\SetKwFunction{ClientPlayAudio}{ClientPlayAudio}

\usepackage{SCITEPRESS}

\begin{document}
\title{AMAVA: Adaptive Motion-Aware Video-to-Audio Framework for Visually-Impaired Assistance}

\author{\authorname{Benjamin Klein\sup{1}\thanks{These authors contributed equally to this work. Published in the Proceedings of the 15th International Conference on Pattern Recognition Applications and Methods (ICPRAM 2026), pages 282--289. DOI: 10.5220/0014289700004067.}, Kazi Ruslan Rahman\sup{2}\sup{a} and Sanchita Ghose\sup{3}}
\affiliation{\sup{1}Department of Computer Science, San Francisco State University, San Francisco, CA 94132, USA}
\affiliation{\sup{2}Department of Mathematics, San Francisco State University, San Francisco, CA 94132, USA}
\affiliation{\sup{3}Department of Computer Engineering, San Francisco State University, San Francisco, CA 94132, USA}
\email{bklein@sfsu.edu, krahman2@mail.sfsu.edu, sanchitaghose@sfsu.edu}
}

\keywords{Accessibility, Assistive Technologies, Real-Time Systems, Sound Effects, Video-to-Audio Generation, Deep Learning, Multimodal Perception.}

\abstract{
    Navigational aids for blind and low vision individuals struggle conveying dynamic real-world environments, leading to cognitive overload from continuous, undifferentiated feedback. We present AMAVA, a novel real-time video-to-audio framework that converts mobile device video into contextually relevant sound effects or text-to-speech descriptions. We propose a motion-aware pipeline using a lightweight AI classification model to distinguish between low and high-movement scenes  followed by a real-time text-to-audio synthesis pipeline to enhance environmental perception more efficiently. In static environments, AMAVA generates spoken audio scene descriptions for situational awareness. In high-movement situations, it prioritizes safety by delivering sound cues, such as spoken hazard alerts and environmental sound effects. These audio outputs are produced by a decoder-only transformer-based vision-language model with mixture-of-experts and cross-modal attention for visual understanding, in conjunction with a neural text-to-speech and natural sound synthesis networks. The proposed framework uses prompt-based caching and category-specific throttling to avoid auditory clutter and minimize latency. We present a comprehensive evaluation of the system, including a real-time navigation study comparing a white cane alone versus with AMAVA, that shows a significant increase in user confidence and perceived safety.
}

\onecolumn \maketitle \normalsize \setcounter{footnote}{0} \vfill
\begin{figure*}[t]
    \centering
    \includegraphics[width=0.9\textwidth]{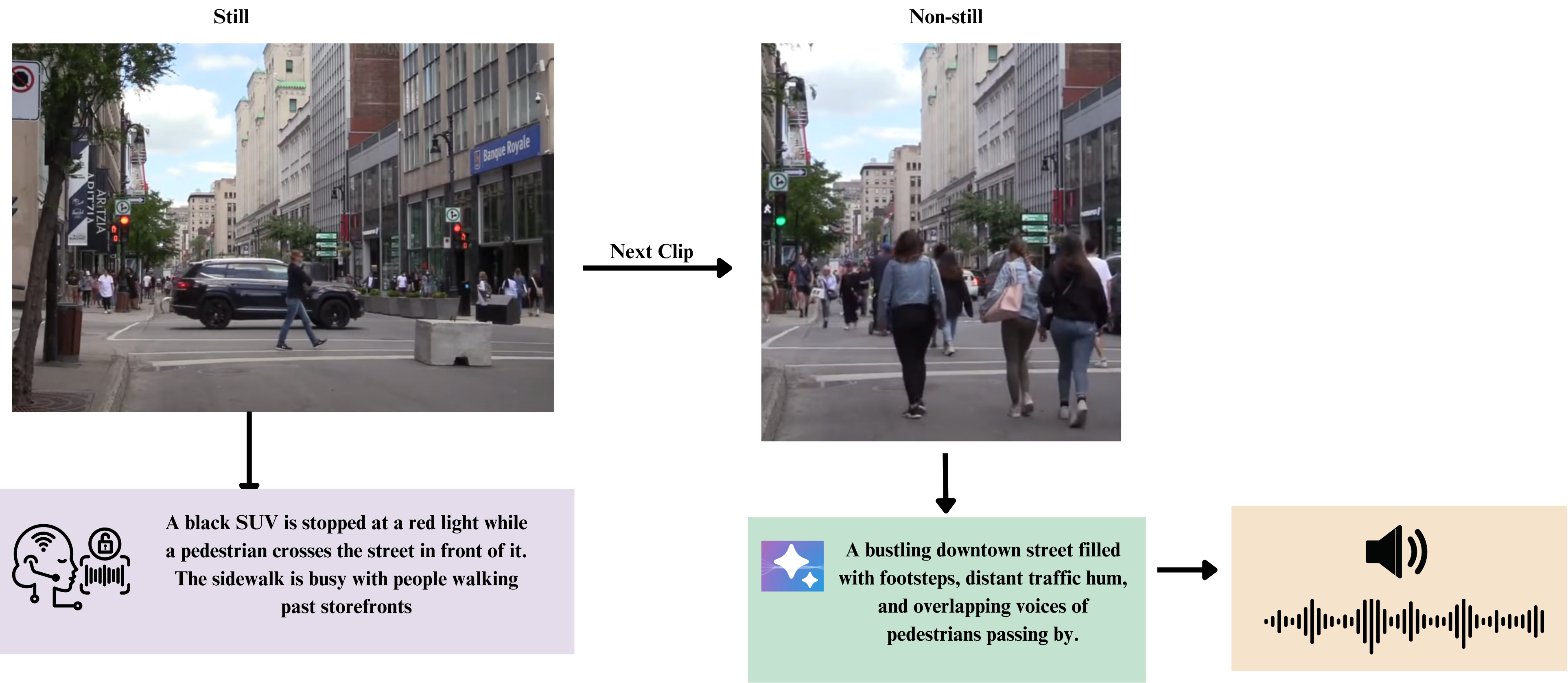}
    \caption{System Overview: Our system aims to provide visually impaired users with audio feedback only when it is most relevant in order to minimize cognitive overload. The left panel shows a still scene, prompting a short descriptive caption; the right panel depicts a non-still, dynamic street, which triggers sound effects or warnings about potential hazards. By distinguishing still from non-still scenes, the system decides when to describe the environment vs. when to generate and play audio cues, offering a balance between essential information and minimal auditory clutter.}
    \label{fig:problem_statement}
\end{figure*}

\section{\uppercase{Introduction}}
\label{sec:introduction}

Navigating complex environments is a challenge for visually impaired individuals. While traditional aids like white canes detect immediate obstacles, they lack broader contextual awareness of pedestrian flow or oncoming vehicles. Audio feedback offers a richer alternative to haptic alerts, yet current tools often fail to meet real-time needs. Applications like Seeing AI~\cite{seeing_ai}, Lookout~\cite{lookout}, and Envision AI~\cite{envision_ai} provide static snapshots rather than continuous updates, while other navigation tools rely on overwhelming text-to-speech (TTS). To address this, we propose AMAVA, a framework designed to reduce cognitive load by distinguishing between low- and high-motion scenes, providing descriptive TTS for static environments and sound effects (SFX) during dynamic moments to avoid overwhelming the user.

Unlike frameworks such as SVA~\cite{sva2024} that process pre-recorded footage using large models~\cite{audiogen}, AMAVA is optimized for real-time mobile use. It processes image batches in parallel and utilizes caching to minimize latency.

Our primary contributions are: (1) A novel real-time accessibility system integrating sound synthesis and TTS; (2) A motion-aware classification system that reduces cognitive overload by distinguishing static from dynamic scenes and throttling audio accordingly; and (3) Evaluations combining system performance metrics with user feedback to demonstrate improved situational awareness.

The rest of this paper is organized as follows: Section II reviews related works in assistive technology. Section III presents an overview of the system architecture, followed by training and implementation details in Section IV. Section V describes our evaluations, which include a navigation test, an ablation test, and a performance test. Finally, Section VI outlines directions for future work.

\section{\uppercase{Related Works}}

Assistive technology has evolved from sensor-based tools to AI-driven applications. Smartphone apps like Seeing AI~\cite{seeing_ai}, Lookout~\cite{lookout}, and Envision AI~\cite{envision_ai} provide on-demand captioning but lack the continuous feedback necessary for safe navigation. Human-in-the-loop services like Aira~\cite{aira} offer useful guidance but are limited by cost and scalability. Wearable aids range from simple ultrasonic sensors~\cite{ultrasonic} to camera-based devices like OrCam~\cite{orcam}, which are often expensive. Research prototypes such as NavCog~\cite{navcog} and CamIO~\cite{camio} rely on prepared environments (e.g., marked with beacons), while SLAM-based approaches~\cite{slam_nav} typically require heavy computational hardware not suitable for standard mobile use. Low-light image enhancement methods such as MSR-MIRNet~\cite{mou2023msrmirnet} improve object visibility in dark environments, but they focus on visual enhancement rather than real-time assistive audio feedback.

In the domain of Video-to-Audio (V2A) generation, our work draws inspiration from the SVA framework~\cite{sva2024}, which pipelines Gemini~\cite{gemini} and AudioGen~\cite{audiogen} to generate semantically accurate audio. However, SVA and similar state-of-the-art models like STA-V2A~\cite{ren_sta-v2a_2025}, Diff-Foley~\cite{luo_diff-foley_2023}, VTA-LDM~\cite{xu_video--audio_2024}, AutoFoley~\cite{ghose2020autofoley}, and FoleyGAN~\cite{ghose2022foleygan} are designed for offline processing of full video clips. To our knowledge, there is a gap in real-time V2A generation. Inspired by SoundEYE~\cite{ghose2024soundeye}, AMAVA addresses this by introducing a branching pipeline that dynamically switches between TTS and SFX. This approach, combined with audio caching, minimizes both cognitive load and latency for live navigation.

\begin{figure*}[t]
    \centering
    \fbox{\includegraphics[width=0.99\textwidth]{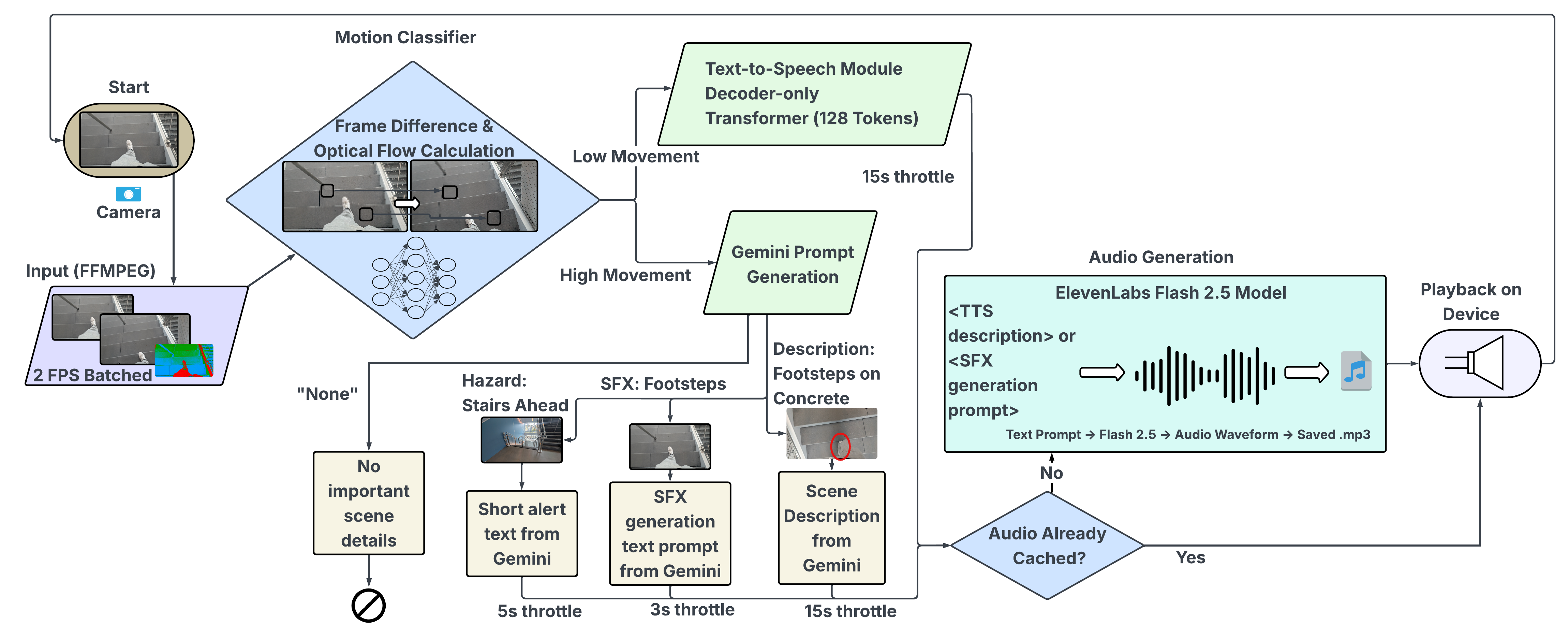}}
    \caption{System overview of the AMAVA pipeline. Camera input is batched and classified into low or high movement, triggering logic for captioning, tagging, and audio generation.}
    \label{fig:system_overview}
\end{figure*}

\section{\uppercase{System Architecture}}

AMAVA has two main components: a browser-based frontend and a Python-based backend server, connected through a WebSocket~\cite{fette2011rfc}. Once the system is activated, the frontend captures frames from the user's camera at a rate of 2 Hz (every 0.5 seconds). These frames are transmitted in real-time to the backend, where they are buffered into 2-frame batches every second. Each batch is processed asynchronously by a movement classifier, determining whether the scene exhibits high or low movement and branching the pipeline into two different audio generation paths. The backend handles captioning, classification, and audio synthesis in parallel to minimize blocking and return low-latency responses. Audio output is streamed back to the frontend for immediate playback. The proposed system workflow is shown in Figure~\ref{fig:system_overview} and the entire AMAVA Video-to-Audio Pipeline is mentioned in Algorithm 1. The following subsections describe the motion classifier pipeline architecture, as well as its operation in analyzing diverse visual scenarios to generate audio.

\begin{algorithm}[!h]
\caption{AMAVA Video-to-Audio Pipeline.}\label{alg:amava}
\KwIn{Live video stream}
\KwOut{Text-to-speech or sound effect audio}

Capture image frames at 2 Hz\;

\ForEach{batch of 2 frames}{
  $movement \gets$ \ClassifyMovement{batch}\;

  \uIf{$movement$ is Low}{
    $response \gets$ \GenerateCaption{batch, ``low''}\;
    $category \gets$ ``description''\;
    \uIf{\IsCached{response}}{
      $audio \gets$ \GetCachedAudio{response}\;
    }\Else{
      $audio \gets$ \TextToSpeech{response}\;
      \CacheAudio{response, audio}\;
    }
  }
  \Else{
    $response \gets$ \GenerateCaption{batch, ``high''}\;
    $(category,\ content) \gets$ \ParseResponse{response}\;
    \uIf{$category$ is none}{\Continue}
    \uIf{\IsCached{response}}{
      $audio \gets$ \GetCachedAudio{response}\;
    }\Else{
      \uIf{$category$ is hazard or description}{
        $audio \gets$ \TextToSpeech{content}\;
      }\ElseIf{$category$ is sfx}{
        $audio \gets$ \GenerateSFX{content}\;
      }
      \CacheAudio{response, audio}\;
    }
  }

  \uIf{\ShouldThrottle{category, response}}{\Continue}
  \SendToClient{audio}\;
  \ClientPlayAudio{audio}\;
}
\end{algorithm}

\subsection{Motion Classifier Pipeline}

To adapt audio output based on user activity, the backend includes a lightweight motion classifier that labels each video segment as either \textit{low} or \textit{high} movement. This binary classification determines which audio logic path is triggered (see Figure~\ref{fig:system_overview}). The classifier was trained on the UCF101 dataset~\cite{ucf101}, using two motion-related features computed for each two-frame batch: \textbf{frame difference} and \textbf{optical flow magnitude}. The first feature, frame difference, is the mean absolute pixel-wise difference ($\bar{D}$) between two consecutive grayscale frames, $I_{t+1}$ and $I_t$:
\begin{equation}
    \bar{D} = \frac{1}{WH} \sum_{x=1}^{W} \sum_{y=1}^{H} |I_{t+1}(x, y) - I_t(x, y)|
\end{equation}

The second feature, optical flow magnitude, is computed using the Farneback algorithm~\cite{farneback}. The mean optical flow magnitude ($\bar{M}$) across the entire frame is calculated as:
\begin{equation}
    \bar{M} = \frac{1}{WH} \sum_{x=1}^{W} \sum_{y=1}^{H} \sqrt{u(x, y)^2 + v(x, y)^2}
\end{equation}
where $u(x, y)$ and $v(x, y)$ are the horizontal and vertical components of the flow vector at each pixel. These two scalar features are then combined to form the 2D input vector $\mathbf{x}$ for the motion classifier:
\begin{equation}
    \mathbf{x} = [\bar{D}, \bar{M}]
\end{equation}

These features capture both static background shifts and localized object motion. Each video batch is pre-processed in real time: the frames are converted to grayscale, the optical flow is computed, and the features are normalized using a prefit \texttt{ standardScaler~\cite{scikit}}. The resulting 2D feature vector is passed to a fully connected neural network (FCNN) implemented in PyTorch.

\subsubsection{Model Architecture}

The FCNN processes the input 2D motion feature vector to output a softmax probability over three movement categories: \textit{low}, \textit{medium}, and \textit{high}. In the current system, the \textit{high} and \textit{medium} predictions are grouped together and treated as a single \textit{high} movement state for downstream logic. The model is small enough for real-time inference and performs reliably on coarse motion classification. The network architecture is as follows:

\begin{itemize}
    \item \textbf{Input layer:} 2D motion feature vector
    \item \textbf{Hidden layer 1:} 32 units, ReLU~\cite{relu} activation
    \item \textbf{Dropout:} 50\% dropout~\cite{dropout} rate to reduce overfitting
    \item \textbf{Hidden layer 2:} 16 units, ReLU activation
    \item \textbf{Output layer:} 3 units, softmax activation (low/medium/high)
\end{itemize}

\subsection{Scene Interpretation and Audio Categorization Pipeline}

We use a transformer-based vision-language model (Gemini~\cite{gemini}) conditioned via prompt engineering to classify two-frame clips into semantic categories: \texttt{sfx}, \texttt{hazard}, \texttt{description}, or \texttt{none}. These categories are used to determine the type of audio feedback generated.

\begin{table*}[t]
    \centering
    \caption{Throttle and Playback Behavior by Audio Type.}
    \begin{tabular}{|l|l|c|l|l|}
        \hline
Category & Audio Type & Throttle & Movement Branch & Playback Condition \\
\hline
        Hazard            & TTS (short)         & 5 sec             & High                     & Always played                                                                            \\
        Sound Effect      & SFX                 & 3 sec             & High                     & Only played if cached                                                                    \\
        Description       & TTS (long)          & 15 sec            & Both                     & Always played                                                                            \\
        None              & —                   & —                 & High                     & No audio                                                                                 \\
\hline
        \multicolumn{5}{l}{\rule{0pt}{2ex}\footnotesize Note: A shared 4-second throttle timer is applied between any two TTS clips (hazard and description) to prevent overlap.} \\
    \end{tabular}
    \label{tab:cache_throttle}
\end{table*}

\subsubsection{Low-Movement Branch}

When movement is minimal, the system delivers ambient scene context. Gemini~\cite{gemini} is prompted to generate a short (15-20 word) scene description based on the visual content of the frame batch. This description is synthesized using ElevenLabs Turbo v2, a sequence-to-sequence TTS model with neural vocoder and emotion-aware intonation. The resulting speech is then played immediately. To avoid redundancy in largely static environments, natural language output is cached with hash-based indexing and throttle-timed reuse. Using caching and enforcing a throttle interval (see Table~\ref{tab:cache_throttle}) maintains relevant feedback during idle moments without being obtrusive.

\subsubsection{High-Movement Branch}

During active navigation or dynamic scenes, the system prioritizes urgency and spatial awareness. The vision-language model uses cross-modal attention to assign the scene to one of four semantic output types: \textit{hazard}, \textit{sound effect}, \textit{description}, or \textit{none}. Hazards are voiced as short alerts. Descriptions offer additional context. Sound effects convey environmental cues but are only played if already cached to reduce latency. Throttle intervals are adjusted per category. See Table~\ref{tab:cache_throttle}.

\subsection{Audio Feedback and Emission Pipeline}

After classification, the backend requests either a TTS caption or SFX clip from the API, which is pretrained on Eleven Flash v2.5 text-to-sound model~\cite{elevenlabs}, based on the Gemini-assigned category. Each response is returned as an MP3 file and streamed to the frontend via WebSocket for playback. Before requesting audio from the API, the system checks a local cache. Each prompt is hashed using SHA-256 after normalization, ensuring consistent reuse and preventing redundant API calls. Because SFX generation introduces delay, sound effects are only played when they are already cached; TTS is played immediately regardless of cache status. All new audio is stored for future reuse. As detailed in Table~\ref{tab:cache_throttle}, throttle intervals for each audio category help manage audio output and reduce clutter. All batch logic runs asynchronously, allowing camera capture and motion classification to proceed in parallel without blocking on audio generation.

\section{\uppercase{Training and Implementation Details}}
\subsection{Motion Classifier Training}

The UCF101-based training set is split into 60\% training, 20\% validation, and 20\% testing subsets. Training of the FCNN is conducted using the cross-entropy loss function and the Adam optimizer~\cite{adam} (learning rate 0.001, weight decay 1e-5). A batch size of 64 ensured stable updates without excessive memory overhead. Early stopping was used to prevent overfitting, halting training if validation loss failed to improve for five consecutive epochs. The model with the lowest validation loss was saved as a .pth file for deployment. During inference, incoming motion features are standardized using the same scaler fitted during training (\texttt{scaler.pkl}) to maintain consistency of features.

\subsection{System Integration}

The AMAVA back-end is implemented in Python using Flask~\cite{flask}, with asynchronous real-time communication handled by Flask-SocketIO~\cite{flask_socketio}. Video frames sent from the web frontend are buffered into batches for processing. The core pipeline operates entirely in-memory and runs in separate threads to maintain responsiveness. This pipeline includes movement classification, a transformer-based multimodal captioning pipeline conditioned on short-term visual context~\cite{gemini}, and audio synthesis using the ElevenLabs API ~\cite{elevenlabs}. For debugging and analysis, key events, performance data, and other metadata are logged to disk. To enable efficient caching, audio prompts are first processed by lowercasing the text and removing punctuation, then hashed. The resulting audio is stored as an MP3 file, indexed by its hash to avoid collisions. These files are returned to the frontend via a WebSocket event, which triggers playback and displays the corresponding caption. System configuration, including batch size and frame rate, is managed through a centralized module. For development and remote testing, the frontend can be exposed using tunneling tools such as \texttt{ngrok}~\cite{ngrok}.

\begin{figure*}[t]
    \centering
    \includegraphics[width=0.65\textwidth]{figures/figure4.png}
    \caption{Heatmap of CLAP cosine similarity scores between generated audio SFX (rows) and their corresponding text prompts (columns). The strong diagonal indicates that the generated audio is most similar to its intended prompt. High off-diagonal values reveal semantic or acoustic confusions.}
    \label{fig:clap_heatmap}
\end{figure*}
\begin{figure*}[t]
    \centering
    \begin{subfigure}[t]{0.32\textwidth}
        \centering
        \includegraphics[width=\textwidth]{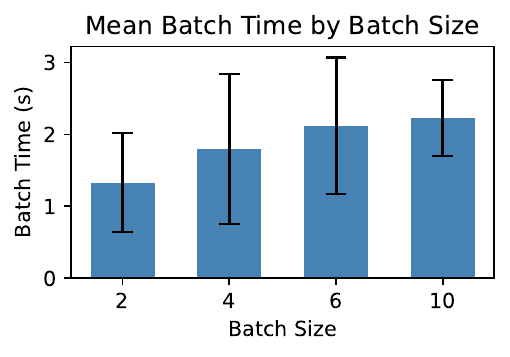}
        \caption{Processing Latency}
        \label{fig:latency}
    \end{subfigure}
    \hfill
    \begin{subfigure}[t]{0.32\textwidth}
        \centering
        \includegraphics[width=\textwidth]{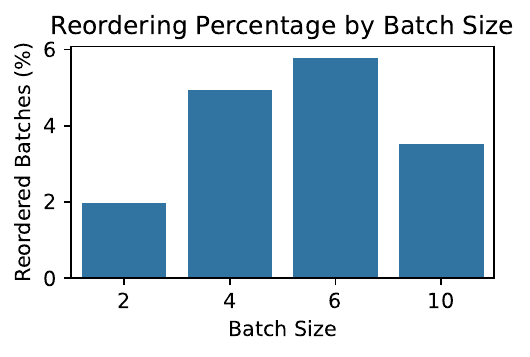}
        \caption{Reordering Rate}
        \label{fig:reordering}
    \end{subfigure}
    \hfill
    \begin{subfigure}[t]{0.32\textwidth}
        \centering
        \includegraphics[width=\textwidth]{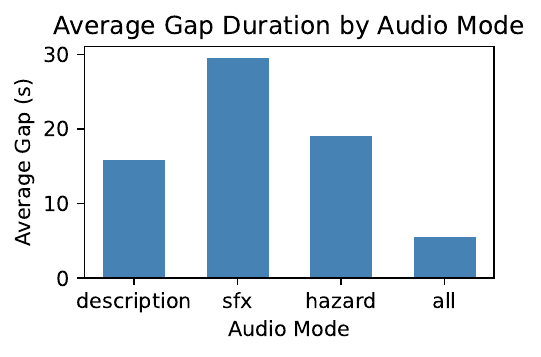}
        \caption{Audio Gap Duration}
        \label{fig:gap}
    \end{subfigure}
    \caption{System performance metrics. (a) Latency increases with batch size, whereas (b) playback reordering peaks at intermediate sizes. (c) The average gap duration indicates the relative frequency of each audio category during the trial.}
    \label{fig:perf_figures}
\end{figure*}

\section{\uppercase{Evaluation}}

\subsection{Motion Classification}
We benchmarked our motion classifier against 40 manually labeled video clips (20 low, 20 high movement) from the UCF101 dataset~\cite{ucf101}. The classifier reached 100\% accuracy in identifying low-motion scenes.For dynamic scenes, it classified all remaining clips as either 'medium' or 'high', both of which trigger the high motion pipeline branch. For the system's intended binary branching, the classifier showed 100\% accuracy in distinguishing static from dynamic environments.

\subsection{Audio Quality (FAD \& CLAP)}
To validate the generated sound effects, we used two metrics: Fréchet Audio Distance (FAD)~\cite{fad} for perceptual quality and CLAP~\cite{clap} for semantic accuracy.
First, using 67 diverse text prompts, we compared our system's output (via ElevenLabs) against real-world recordings from Freesound.org~\cite{freesound}. As shown in Table~\ref{tab:fad_comparison}, the ElevenLabs sound effects reached an FAD score of 1.71, outperforming state-of-the-art baselines like AudioLDM2 (2.85) and AudioGen (3.01).
Second, we evaluated semantic alignment using a 10x10 CLAP similarity matrix on generated clips (e.g., "keyboard typing", "coughing"). The results showed a high semantic accuracy, with correct audio-text pairs a mean cosine similarity of 0.32, compared to -0.02 for incorrect pairs. These results show that the generated audio is semantically accurate to the prompts.

\begin{table}[h]
    \centering
    \caption{Fréchet Audio Distance (FAD) vs. Real Recordings.}
    \label{tab:fad_comparison}
    \small
    \begin{tabular}{l c}
        \toprule
        Model & FAD $\downarrow$ \\
        \midrule
        AudioLDM large~\cite{liu2023audioldm} & 3.12 \\
        AudioLDM small v2 & 3.06 \\
        AudioGen-medium & 3.01 \\
        AudioLDM2~\cite{liu2024audioldm2} & 2.85 \\
        \textbf{AMAVA (Using Elevenlabs)} & \textbf{1.71} \\
        \bottomrule
    \end{tabular}
\end{table}

\subsection{System Performance}

Performance was analyzed using logs from a 7-minute outdoor walk. As illustrated in Figure~\ref{fig:perf_figures}a, mean processing time increased with batch size, ranging from 1.3s (2 frames) to 2.2s (10 frames). Larger batches caused varied processing times, causing audio cues to play out of order (Figure~\ref{fig:perf_figures}b). To minimize this problem, a batch size of 2 was selected for user studies. Figure~\ref{fig:perf_figures}c shows that our throttling successfully paced the audio feedback, with an average interval of less than 5 seconds between audio events.

\subsection{User Study \& Ablation}
We conducted a study with 12 sighted volunteers (blindfolded to simulate low vision) navigating indoor and outdoor obstacle courses under two conditions: White Cane only (Control) vs. White Cane + AMAVA.
\subsubsection{Results} While completion times were similar (approx. 3m 12s for both), feedback was highly positive (Table~\ref{tab:survey_results}). Participants reported significantly higher perceived safety (4.8/5) with AMAVA.
\subsubsection{Ablation} To test the importance of the motion classifier, we tested the full system against a "static-only" version (TTS descriptions only). The full system reduced navigation time (119s vs. 126s) and was preferred by users (3.43/5) over the ablation version.

\begin{table}[h]
    \centering
    \caption{Full Survey Results (Mean $\pm$ SD). Negatively framed questions are reverse-scored (Higher is better).}
    \label{tab:survey_results}
    \small
    \begin{tabular}{l c}
        \toprule
        \textbf{Metric (Abbreviated)} & \textbf{Score} \\
        \midrule
        \textit{System Feedback} & \\
        Helpful & $4.20 \pm 0.45$ \\
        Accurate Alerts & $4.00 \pm 0.71$ \\
        Timely Alerts & $3.60 \pm 1.14$ \\
        Perceived Safety (vs. Cane) & $4.80 \pm 0.45$ \\
        Not Distracting & $4.20 \pm 0.84$ \\
        \midrule
        \textit{Usability (SUS)} & \\
        Low Complexity & $4.20 \pm 0.84$ \\
        Easy to Use & $4.40 \pm 0.89$ \\
        No Tech Support Needed & $4.40 \pm 0.89$ \\
        Well Integrated & $4.00 \pm 1.00$ \\
        High Consistency & $3.60 \pm 0.89$ \\
        Quick to Learn & $4.80 \pm 0.45$ \\
        Not Cumbersome & $3.40 \pm 1.67$ \\
        Confident Using System & $4.00 \pm 0.71$ \\
        No Prior Learning Req. & $5.00 \pm 0.00$ \\
        \midrule
        \textit{Ablation Study} & \\
        Not Overwhelming & $3.85 \pm 1.06$ \\
        Easier than Static Mode & $3.43 \pm 0.98$ \\
        Prefer Full System & $3.43 \pm 0.98$ \\
        \bottomrule
    \end{tabular}
\end{table}

\subsection{Identified Limitations}
Current limitations include the lack of spatial audio (making directionality difficult), the 1-second processing latency, and reliance on cloud APIs which depend on connectivity. Future work will focus on integrating spatial audio cues and on-device processing to address these gaps.

\section{\uppercase{Conclusion and Future Work}}

This research presents AMAVA, a real-time, motion-aware video-to-audio framework designed to support blind and low-vision individuals during navigation in dynamic environments. The system reduces cognitive overload by adjusting audio feedback based on motion. It gives spoken descriptions in still scenes and alerts or sound effects when movement is high. AMAVA combines AI-based motion classification with generative audio, using techniques such as prompt-based caching and category-specific throttling to manage latency and feedback quality in real-time. We conduct a preliminary user study comparing the use of a white cane alone versus a cane with AMAVA. While task completion times were similar across both conditions, participants reported a notable increase in confidence and perceived safety when AMAVA was used. The system is also described as easy to learn and use. Future extensions could explore a direct video-to-audio pipeline to reduce latency, the addition of spatial audio using head-related transfer functions (HRTFs~\cite{hrtf1993}) for better directional awareness, and persistent scene understanding through simultaneous localization and mapping (SLAM~\cite{slam}). User experience may also benefit from additional settings, like adjustable feedback verbosity. A larger-scale study would help further assess real-world impact.

\section*{\uppercase{Acknowledgments}}

This research is sponsored by the National Science Foundation (Award Number: 2347727). The authors would like to thank the AI-LAMP research lab team for the navigation experiment and data collection. Special gratitude is extended to Dr. Sanchita Ghose for her expert supervision, insightful feedback, and unwavering encouragement throughout this project.

\bibliographystyle{apalike}
{\small
    \bibliography{refs}}

\end{document}